\documentclass[11pt,a4paper]{article}
\usepackage[hyperref]{acl2021}
\usepackage{times}
\usepackage{latexsym}

\usepackage{microtype}
\usepackage{multirow}

\usepackage{amsthm,amsmath,amssymb}
\usepackage{mathrsfs,mathtools}

\title{Automatic Biomedical Term Clustering by \\
Learning Fine-grained Term Representations}

\newcommand*\samethanks[1][\value{footnote}]{\footnotemark[#1]}

 \author{
Sihang Zeng \thanks{$\quad$Contributed equally.} \space\space
Zheng Yuan \samethanks \space\space\space
Sheng Yu \thanks{$\quad$Corresponded author.} \\
Tsinghua University \\
\texttt{\{zengsh19,yuanz17\}@mails.tsinghua.edu.cn} \\
\texttt{syu@tsinghua.edu.cn}
}

\date{}

\begin{document}
\maketitle
\begin{abstract}
Term clustering is important in biomedical knowledge graph construction.
Using similarities between terms embedding is helpful for term clustering.
State-of-the-art term embeddings leverage pretrained language models to encode terms, and use synonyms and relation knowledge from knowledge graphs to guide contrastive learning.
These embeddings provide close embeddings for terms belonging to the same concept.
However, from our probing experiments, these embeddings are not sensitive to minor textual differences which leads to failure for biomedical term clustering.
To alleviate this problem, we adjust the sampling strategy in pretraining term embeddings by providing dynamic hard positive and negative samples during contrastive learning to learn fine-grained representations which result in better biomedical term clustering.
We name our proposed method as CODER++\footnote{Our codes and model will be released at \url{https://github.com/GanjinZero/CODER}.},
and it has been applied in clustering biomedical concepts in the newly released Biomedical Knowledge Graph named BIOS\footnote{\url{https://bios.idea.edu.cn/}}.
\end{abstract}

\section{Introduction}
A critical step for building a biomedical knowledge graph is clustering synonyms terms into concepts \cite{nicholson2020constructing,bios}.
After mining terms from the biomedical corpus or electronic medical records, these terms may belong to an existing concept dictionary or newly discovered concepts.
It is hard for humans to link terms to an existing concept dictionary since the volume of the concept dictionary is huge.
Furthermore, it is almost impossible for humans to determine if one term is a new concept. 

Embedding-based entity linking methods encode terms into a dense space and use similarities among terms for entity linking \cite{sapbert,coder}.
Terms that belong to newly discovered concepts should have low similarities to all concepts in the dictionary.
Embedding-based entity linking methods can also assist humans in term clustering by providing candidates.
However, we find that existing state-of-the-art biomedical term embedding models SapBERT \cite{sapbert} and CODER \cite{coder} are not sensitive to fine-grained differences (i.e. They provide high similarities for non-synonymous and textually similar term pairs).
These term pairs are common, especially in diseases (e.g. \textit{Type 1 Diabetes} v.s. \textit{Type 2 Diabetes}) and chemicals (e.g. \textit{xyloglucan endotransglycosylase} v.s. \textit{xyloglucan endoglucanase}).
We suggest the reason comes from the pretraining sampling strategy of SapBERT and CODER.
They sample Concept Unique Identifiers (CUIs) from UMLS \cite{umls} randomly in the mini-batch.
This produces hard positive pairs (i.e. textually different terms with the same CUIs) and easy negative pairs (i.e. textually different terms with different CUIs).
Supervised contrastive learning is applied to cluster embeddings under the same CUIs and to keep away embeddings for different CUIs.
For benchmarking entity linking tasks, the ability to determine positive pairs is important. For term clustering, it further requests to determine negative pairs. 
Hard negative pairs are absent in pretraining SapBERT and CODER which lead to unsatisfactory performances in term clustering.

In this paper, we propose a probing experiment to evaluate term clustering on UMLS automatically. 
This experiment shows SapBERT and CODER have insufficient ability in term clustering.
For better term clustering, we propose a dynamic sampling strategy that provides both hard positive and negative pairs to learn fine-grained terms embeddings named CODER++.
CODER++ not only reserves the ability to normalize terms but also can distinguish different concepts with similar term names.
CODER++ shows decent ability on biomedical entity linking and a significant improvement on biomedical term clustering evaluation.


\section{Related Work}
Automatic term clustering has long been discussed.
Traditional methods use statistical approaches to define similar terms and perform clustering. \citet{automaticretrieval} defines term similarity based on distributions and \citet{2001syntacticphrases} forms clusters based on co-occurrence in semantically coherent documents.
\citet{kok2008extracting} uses Markov logic for unsupervised concept clustering.

Recent researches focus on deep learning approaches, where biomedical term embeddings can be used for term clustering. \citet{nguyen2015identifying} identifies biomedical synonyms using word embeddings. SapBERT \cite{sapbert,umlseval} and CODER \cite{coder} learn synonyms knowledge from UMLS to provide close embeddings for synonyms. In this work, we improve these embeddings by providing dynamic hard negative samples.









\begin{table*}[t]
\small
\centering
\begin{tabular}{cccccc}
\hline
\multirow{2}{*}{Term 1}               & \multirow{2}{*}{Term 2}        & \multicolumn{3}{c}{Similarity}                    & \multirow{2}{*}{Same CUI} \\ \cline{3-5}
                                      &                                & CODER & SapBERT & CODER++                         &                           \\ \hline
julibroside j2                        & julibroside c1                 & 0.918 & 0.918   & \textbf{0.339} & F                         \\
orange colored urine                  & pink urine                     & 0.738 & 0.783   & \textbf{0.451} & F                         \\
type 2 diabetes 1                     & type 1 diabetes                & 0.908 & 0.911   & \textbf{0.502} & F                         \\
sb 212047                             & sb 216754                      & 0.819 & 0.767   & \textbf{0.356} & F                         \\
early onset                           & late onset                     & 0.831 & 0.807   & \textbf{0.416} & F                         \\
ginsenoside rh                        & ginsenoside rg                 & 0.908 & 0.979   & \textbf{0.420} & F                         \\
protein phosphatase 1 delta           & protein phosphatase 2c delta   & 0.910 & 0.832   & \textbf{0.616} & F                         \\
type ii endometrial carcinoma         & endometrial cancer stage ii    & 0.845 & 0.846   & \textbf{0.420} & F                         \\ \hline
headache                              & cephalgia                      & 0.798 & 0.741   & 0.776                           & T                         \\
fhx allergies                         & fh: allergy                    & 0.879 & 0.881   & 0.819                           & T                         \\
herpesvirus murid 004                 & murine herpesvirus 068         & 0.634 & 0.823   & 0.674                           & T                         \\
tex2                                  & tex2 gene                      & 0.890 & 0.995   & 0.921                           & T                         \\
eppin 1 protein, human                & eppin protein, human           & 0.991 & 0.941   & 0.834                           & T                         \\
chmp2b gene                           & chromatin modifying protein 2b & 0.743 & 0.797   & 0.724                           & T                         \\ \hline
\end{tabular}
\caption{Similarities of different models between representative term pairs with the same CUI or different CUI. Term pairs with the same CUI are considered positive. Compared with CODER and SapBERT, CODER++ has relatively lower similarities on negative term pairs and moderately higher similarities on positive term pairs.}
\label{tab:fp coder}
\end{table*}

\section{Term Clustering Evaluation}
We introduce the term clustering evaluation on UMLS as the probing experiment, in which we find that 
both CODER and SapBERT show poor clustering performance. Through the case study, we find the reason is that both models fail to distinguish between fine-grained biomedical terms, which suggests a refinement is needed to support biomedical term clustering.


\subsection{Embedding-based Term Clustering}
\label{umls cluster}
We use term embeddings including CODER and SapBERT to perform clustering on UMLS terms. 
We first generate embedding $e$ for each term $t$ in UMLS. 
The similarity between term $t_i$ and $t_j$ is measured by cosine similarity $S_{ij}=cosine(e_i, e_j)$.
If $S_{ij}>\theta$, where $\theta$ is a hyperparameter, $t_i$ and $t_j$ are predicted to be clustered.
In practice, calculating similarities between all pairs is time-consuming.
Instead, for each term $t_i$, we use the Faiss index \cite{faiss} to only save terms with top-$m$ similarities with $t_i$, denoted by $\mathcal{M}_i$.
Only when $t_j\in \mathcal{M}_i$ and also $S_{ij}>\theta$, $t_j$ is predicted to be clustered with $t_i$ (i.e. $t_i$ and $t_j$ are synonyms).
For convenience, we denote $\mathcal{M}=\bigcup_i \mathcal{M}_i$.

\subsection{Large-scale Clustering Evaluation}
\label{eval}
For evaluation, terms under the same CUI $i$ in UMLS are regarded as ground truth clustering, denoted by $\mathcal{C}_i$. We denote $\mathcal{C}=\bigcup_i\mathcal{C}_i$. Suppose there are $n$ terms, then we have $\binom{n}{2}$ term pairs. For each pair $(t_i, t_j)$, if they are under the same CUI and also predicted to be clustered, then $(t_i, t_j)$ is regarded as true positive (TP). False positive (FP), false negative (FN), and true negative (TN) are defined similarly.
Recall, precision and $F_1$ score can be computed based on TP, FP, FN, and TN.
Precision suggests how well a model differentiates between negative term pairs. 
Recall suggests how well a model clusters terms with similar meanings. 

As $n$ is large in practice (over 10M terms in a biomedical terminology like UMLS), it is impossible to enumerate all term pairs to directly count TP, FP, FN, and TN.
\citet{umlseval} downsamples negative pairs for evaluation, but this may ignore some hard negative pairs.
We propose an efficient algorithm for large-scale clustering evaluation, which reduces the time complexity from $\mathcal{O}(n^2)$ to $\mathcal{O}(n)$ when ground truth cluster $\mathcal{C}_i$ is bounded. The algorithm splits the searching space into two parts, traversing through the Faiss index $\mathcal{M}$ and traversing through the ground truth cluster space $\mathcal{C}$. When traversing through $\mathcal{M}$, we first get pairs with predicted labels to be true, then count how many pairs in $\mathcal{C}$ to obtain TP and FP. When traversing through $\mathcal{C}$, we first get pairs with ground truth label to be true, then count pairs in $\mathcal{M}$ to obtain FN. TN is computed by subtracting TP, FP, and FN from $\binom{n}{2}$ instead of counting which saves time significantly.
To speed up the searching process, we also store $\mathcal{C}$ and $\mathcal{M}$ in prefix trees.

\subsection{Probing Results}

The results of term clustering evaluation in UMLS 2020 AA for CODER and SapBERT are shown in Table \ref{tab:umls cluster}. 
We search for the best threshold $\theta_0$ according to the $F_1$ score.
$F_1$ scores are both low for SapBERT and CODER, which indicates that both models could not differentiate terms well and tend to cluster different terms together. These $F_1$ scores are much lower than reported in \cite{umlseval} (0.65 for SapBERT), the reason is they downsample negative pairs in evaluation which underestimates FN.
The performance gap between SapBERT and CODER comes from their different sampling strategies.

\begin{table}[t]
\small
\centering
\begin{tabular}{llccc}
\hline
Model   & $\theta_0$ & $P$   & $R$   & $F_1$ \\ \hline
SapBERT & 0.94        & 0.302 & 0.268 & 0.284 \\
CODER   & 0.86        & 0.071 & 0.401 & 0.121 \\
\hline
\end{tabular}
\caption{Results for CODER and SapBERT on term clustering evaluation in UMLS 2020 AA.}
\label{tab:umls cluster}
\end{table}

\subsection{Case Study}
\label{case}
We sample term pairs to check why CODER and SapBERT fail on term clustering evaluation.
Similarities of representative false positive term pairs for both CODER and SapBERT are shown in the upper part of Table \ref{tab:fp coder}. We can observe that CODER and SapBERT embeddings can't distinguish terms with number differences, body part differences, and devices differences.
CODER and SapBERT provide similarities for these false positive term pairs as high as true positive term pairs shown in the lower part of Table \ref{tab:fp coder}.
Hence they tend to cluster terms with highly similar strings but different meanings. 


\section{Approach}
We introduce CODER++ to address the above-mentioned problem.
The idea is simple, providing hard negative pairs to reduce false positive term pairs.
We focus on how to construct mini-batches to learn fine-grained term representations. 

\subsection{Term Encoding}
CODER++ embeds a term $s$ to a dense representation $\mathbf{e}$ with a pretrained language model.
We tokenize $s$ into sub-words, and use the representation of {\rm [CLS]} token for term representation.

\subsection{Dynamic Sampling}
\label{method}
\paragraph{Positive Sampling}
For each term $t$, we sample $k$ terms $p_1,...,p_k$ with same CUI from UMLS.
This adds positive pairs for training. The term $p_i$ can be textually different from $t$ which is considered a hard positive sample.

\paragraph{Possibly Hard Negative Sampling}
We take terms $n_1,...,n_m$ with top-$m$ similarities with term $t$ as possibly hard negative samples.
It is expensive to find terms with top-$m$ similarities on the fly, and we use the Faiss index instead.
For each epoch, we update the Faiss index using the present CODER++. 
Selected terms can have the same CUI or different CUIs with term $t$.
A not well-trained model has more different CUIs terms as hard negative samples.
The model is required to distinguish these fine-grained terms.
When the training is progressed, more selected terms will have the same CUI with the term $t$.

\begin{figure}[t]
    \centering
    \includegraphics[width=\linewidth]{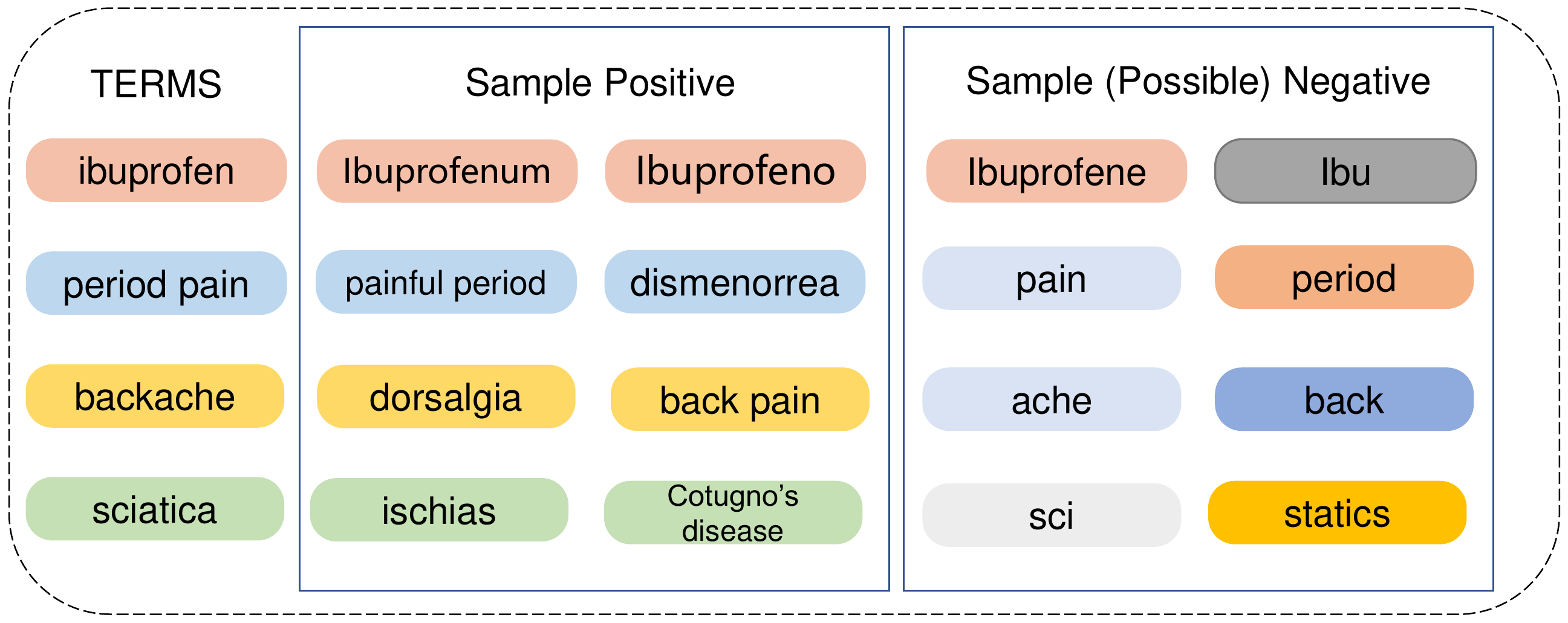}
    \caption{Construction of a mini-batch in CODER++.}
    \label{fig:workflow}
\end{figure}

\paragraph{Overall Sample Strategy}
We first sample terms $\{t_i\}_i$ randomly from the whole term set.
For each term $t_i$, we sample $k$ positive terms $p_{i_1},...,p_{i_k}$ and $m$ possibly hard negative terms $n_{i_1},...,n_{i_m}$.
All these terms $\{t_i,p_{i_1},...,p_{i_k},n_{i_1},...,n_{i_m}\}_i$ construct a mini-batch, and we use the CUIs of these terms to guide supervised contrastive learning. An example of mini-batch is visualized in Figure~\ref{fig:workflow}.
We follow \citet{sapbert,coder} to optimize the model using the Multi-Similarity loss (MS-loss) \citep{wang2019multi} to guide terms with same CUIs similar and terms with different CUIs dissimilar.
\begin{align}
    \mathscr{N}_i&\coloneqq\{j\vert1\leq j\leq m,c_i\neq c_j,S_{ij}>\min_{c_k=c_i}S_{ik}-\epsilon\} \nonumber\\
    \mathscr{P}_i&\coloneqq\{j\vert1\leq j\leq m,c_i=c_j,S_{ij}<\max_{c_k\neq c_i}S_{ik}+\epsilon\} \nonumber\\
    \mathcal{L} &= \frac{1}{m}\sum_{i=1}^m(\frac{\log(1+\sum_{j\in\mathscr{P}_i}\exp(-\alpha(S_{ij}-\lambda)))}{\alpha} \nonumber\\
    &+\frac{\log(1+\sum_{j\in\mathscr{N}_i}\exp(\beta(S_{ij}-\lambda)))}{\beta}), \nonumber
\end{align}
where $c_i$ is the CUI of $i^{th}$ term, and $\epsilon, \alpha, \beta$ are hyperparameters.

\section{Experiments}
\subsection{Pre-training}
We train CODER++ initialized by CODER with 1,200K training steps \footnote{SapBERT can also be used as the initial checkpoint.}. We update the Faiss index every 60K steps. 
For each mini-batch, we set $k=m=30$. 
Training costs 9 days on 8 NVIDIA A100 40GB GPUs. 
Each GPU samples 16 terms $\{t_i\}$ from UMLS 2020 AA at one time with 8 gradient accumulation steps which indicates a total of $16\times(1+30+30)\times8\times8=62,464$ terms for each parameter update step. 
The maximal term length is set to 32. 
We use AdamW \cite{adamw} as the optimizer with a linear warm-up in the first 10000 steps to a peak of 4e-5 learning rate and a linear decay.
The setting of hyperparameters $\epsilon, \alpha, \beta$ in MS-loss is following \cite{coder}.

\subsection{Term Clustering Evaluation}
\begin{figure}[t]
    \centering
    \includegraphics[width=\linewidth]{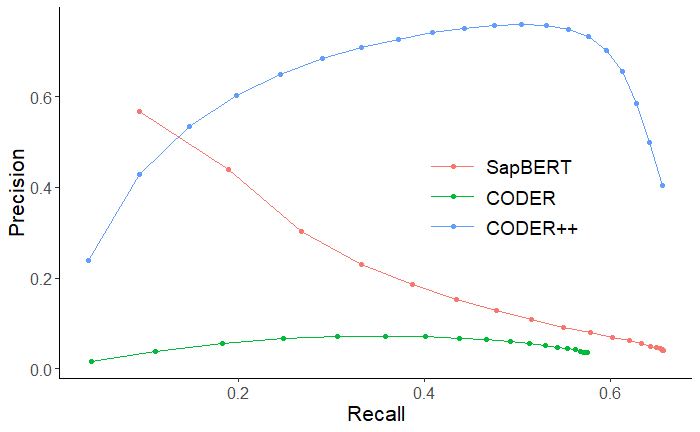}
    \caption{UMLS term clustering evaluation for CODER, SapBERT, and CODER++ under different thresholds.}
    \label{fig:umls}
\end{figure}

We evaluate CODER++ based on Section \ref{eval}. The result is shown in Figure \ref{fig:umls} and Table \ref{tab:umls cluster:pp}. Table \ref{tab:umls cluster:pp} shows that CODER++ greatly outperforms CODER and SapBERT, obtaining 0.732, 0.576, and 0.644 for precision, recall, and $F_1$ scores respectively. We can see from Figure \ref{fig:umls} that CODER++ has a comparable spread in recall with both CODER and SapBERT, which indicates CODER++ reserves the ability of clustering terms with similar meanings, while achieving much better precision for most thresholds, which indicates a significant improvement in distinguishing terms with different meanings.

\subsection{Case Study}
We compute similarities for the same term pairs as in Section \ref{case} using CODER++, and the results are shown in the upper part of Table \ref{tab:fp coder}. It suggests that CODER++ has relatively low similarities on negative term pairs and reduces the FP rate. 
To check if CODER++ maintains high similarities for positive term pairs, we sample some positive terms pairs and compute the similarities, which are shown in the lower part of Table \ref{tab:fp coder}. We observe that CODER++ has moderately high similarities for positive term pairs, which suggests CODER++ reserves the ability to normalize terms with similar meanings.

In conclusion, our dynamic sampling strategy significantly decreases similarities in negative term pairs, while mildly decreasing similarities in positive pairs. The results indicate the efficacy of our dynamic sampling strategy in pretraining.

\begin{table}[t]
\small
\centering
\begin{tabular}{llccc}
\hline
Model   & $\theta_0$ & $P$   & $R$   & $F_1$ \\ \hline
SapBERT & 0.94        & 0.302 & 0.268 & 0.284 \\
CODER   & 0.86        & 0.071 & 0.401 & 0.121 \\
\hline
CODER++ & 0.70        & \textbf{0.732} & \textbf{0.576} & \textbf{0.644} \\ \hline
\end{tabular}
\caption{Results for CODER, SapBERT, and CODER++ on term clustering evaluation in UMLS 2020 AA.}
\label{tab:umls cluster:pp}
\end{table}

\subsection{Zero-shot Term Normalization}
We evaluate CODER++ with zero-shot term normalization on BC5CDR \cite{bc5cdr}, results are shown in Table~\ref{tab:mel}. CODER++ achieves better performance than CODER and comparable performance with SapBERT, which shows that CODER++ generalizes well and reserves the ability to normalize terms with different names.

\begin{table}[t]
\small
\centering
\begin{tabular}{lcc}
\hline
Model & BC5CDR-d   & BC5CDR-c   \\ \hline
SapBERT       & 93.5, 96.0 & 96.5, 98.2 \\
CODER         & 92.2, 94.7 & 95.1, 97.2 \\
CODER++   & 92.2, 94.9 & 96.5, 97.9 \\ \hline
\end{tabular}
\caption{Acc@1 and Acc@5 on BC5CDR for CODER, SapBERT, and CODER++.}
\label{tab:mel}
\end{table}

\subsection{Ablation Study}
Here we conduct ablation studies on sampling strategies. 
Ablation studies are based on a sampled subset of UMLS, which consists of 500K terms (denote as $\mathcal{D}_s$).
We train models with different settings on $\mathcal{D}_s$ respectively, then use each model to perform clustering evaluation on it:

\noindent Setting (a): $k=1, m=30$, do not update Faiss.

\noindent Setting (b): $k=m=30$, do not update Faiss.

\noindent Setting (c): $k=m=30$, update Faiss index every epoch (i.e. proposed CODER++).

\begin{table}[t]
\small
\centering
\begin{tabular}{llcccc}
\hline
Setting&$\theta_0$  & P & R & $F_1$ \\
\hline
CODER&0.88&0.273&0.310&0.290\\
(a)& 0.76&0.482&0.289&0.361\\
(b)&0.74&0.667&0.517&0.583\\
(c)&0.68    & 0.830  & 0.659  & \textbf{0.735}\\
\hline
\end{tabular}
\caption{Ablation study on sampling strategies with $\mathcal{D}_s$ term clustering.}
\label{tab:ablation}
\end{table}

\begin{figure}[t]
    \centering
    \includegraphics[width=\linewidth]{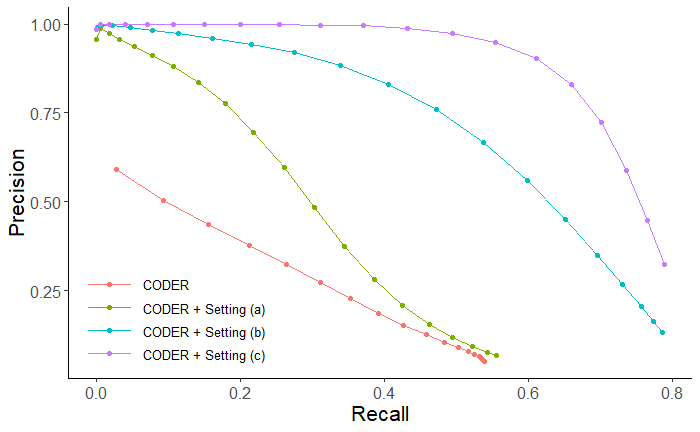}
    \caption{Ablation study on sampling strategies with $\mathcal{D}_s$ term clustering under different thresholds.}
    \label{fig:ablation}
\end{figure}

Figure~\ref{fig:ablation} displays results for thresholds ranging from 0.6 to 0.98, and Table~\ref{tab:ablation} lists the best performances among those thresholds of each model. 
Setting (a) has much higher precision than the original CODER in all thresholds, which indicates hard negative samples do improve the ability to differentiate negative term pairs. 
Setting (b) has higher precision and recall than setting (a), especially recall, which indicates simultaneously using positive and negative samples reserves the ability of clustering similar terms while achieving a better capability of differentiating terms. 
Setting (c) has higher precision than setting (b), which indicates dynamic negative samples greatly enhance the ability to differentiate negative term pairs. The negative sampling under setting (b) is static, the model can easily overfit these samples; while setting (c) will provide new hard negative samples based on the current model.
The result is quite intuitive since dynamic negative samples improve precision and recall simultaneously along with all thresholds.
In conclusion, dynamic negative sampling with balanced positive sampling is the setting that performs best and we use it for training CODER++.

\section{Conclusions}
We propose CODER++, a fine-grained biomedical term representation, which benefits from our dynamic sampling strategy that provides hard positive and negative pairs.
We propose an automatic large-scale clustering evaluation algorithm.
Through a combination of automatic evaluation and the case study, we find CODER++ greatly outperforms CODER and SapBERT on UMLS term clustering and has a much better ability to distinguish different concepts with similar term names.
The effectiveness of our dynamic sampling strategy is also proved through an ablation study.
Our work can be used for automatic term clustering or recommend candidate similar terms for experts and crowdsourcing participants in human term clustering.
Our work also suggests that biomedical term embedding models such as CODER can be further pretrained by focusing on specific information.
\section*{Acknowledgments}
This work was supported by the National Natural Science Foundation of China (Grant No. 12171270), the Natural Science Foundation of Beijing Municipality (Grant No. Z190024), and the International Digitial Economy Academy.


\bibliographystyle{acl_natbib}
\bibliography{acl2021}


\end{document}